\documentclass[lettersize,journal]{IEEEtran}
\usepackage{amsmath,amsfonts}
\usepackage{algorithmic}
\usepackage{algorithm}
\usepackage{array}
\usepackage[caption=false,font=normalsize,labelfont=sf,textfont=sf]{subfig}
\usepackage{textcomp}
\usepackage{stfloats}
\usepackage{url}
\usepackage{verbatim}
\usepackage{graphicx}
\usepackage{cite}
\usepackage{multirow}
\usepackage{bbding}
\usepackage{setspace}
\usepackage{booktabs}
\usepackage{xcolor}
\definecolor{firstcolor}{RGB}{220, 20, 60}    
\definecolor{secondcolor}{RGB}{30, 144, 255}  
\newcommand{\first}[1]{\textcolor{firstcolor}{\textbf{#1}}}
\newcommand{\second}[1]{\textcolor{secondcolor}{\textbf{#1}}}

\begin{document}

\title{Online Handwritten Signature Verification Based on Temporal-Spatial Graph Attention Transformer}
\author{Hai-jie Yuan, Heng Zhang\IEEEauthorrefmark{1}, and Fei Yin

\thanks{
Hai-Jie Yuan is with the State Key Laboratory of Multimodal Artificial Intelligence Systems (MAIS), Institute of Automation, Chinese Academy of Sciences, Beijing 100190, China, and the School of Artificial Intelligence, University of Chinese Academy of Sciences, Beijing 100049, China (e-mail: yuanhaijie2024@ia.ac.cn).

Heng Zhang (Corresponding author) and Fei Yin are with the State Key Laboratory of Multimodal Artificial Intelligence Systems (MAIS), Institute of Automation, Chinese Academy of Sciences, Beijing 100190, China (e-mail: heng.zhang@ia.ac.cn; fyin@nlpr.ia.ac.cn).

}
}


\maketitle

\begin{abstract}
Handwritten signature verification is a crucial aspect of identity authentication, with applications in various domains such as finance and e-commerce. However, achieving high accuracy in signature verification remains challenging due to intra-user variability and the risk of forgery. This paper introduces a novel approach for dynamic signature verification: the Temporal-Spatial Graph Attention Transformer (TS-GATR). TS-GATR combines the Graph Attention Network (GAT) and the Gated Recurrent Unit (GRU) to model both spatial and temporal dependencies in signature data. TS-GATR enhances verification performance by representing signatures as graphs, where each node captures dynamic features (e.g. position, velocity, pressure), and by using attention mechanisms to model their complex relationships. The proposed method further employs a Dual-Graph Attention Transformer (DGATR) module, which utilizes k-step and k-nearest neighbor adjacency graphs to model local and global spatial features, respectively. To capture long-term temporal dependencies, the model integrates GRU, thereby enhancing its ability to learn dynamic features during signature verification. Comprehensive experiments conducted on benchmark datasets such as MSDS and DeepSignDB show that TS-GATR surpasses current state-of-the-art approaches, consistently achieving lower Equal Error Rates (EER) across various scenarios.
\end{abstract} 

\section{Introduction}
\label{sec:intro}

\IEEEPARstart{I}{n} the modern information society, driven by the rapid development of digital technologies, network security and identity verification have become critical concerns in contemporary society\cite{Wang2021Attacks, Tran2022PrivacyPreserving}. Identity verification serves as a key measure to ensure information security and prevent identity theft. It is widely applied in Internet services such as e-commerce, online payments, e-government, and digital banking \cite{Bhavani2024,Rukhiran2023IoT,Wang2025Fingerprinting,Marta2023privacy,Ashbourn2014Biometrics}. Traditional authentication methods, such as passwords and physical ID cards, offer fundamental identity assurance. However, as cyber-attack techniques evolve, these methods have been found to exhibit numerous security vulnerabilities \cite{Tran2022PrivacyPreserving, Wang2021Attacks, Wu2023Attacks,Marta2023privacy,Ashbourn2014Biometrics,Kaur2023}. Therefore, developing more secure, convenient, and reliable identity verification technologies has become imperative.

Handwritten signature verification, a traditional method of personal authentication, has been used for thousands of years. Compared to physiological biometrics such as facial recognition, iris scanning, or fingerprint analysis, handwritten signatures—a form of behavioral biometrics—are easier to capture and more widely accepted by users\cite{hafemann2017offline,blanco2014performance,sharma2022behavioral,buckley2019language}. This has led to their prevalence in administrative, financial, and commercial domains \cite{Wu2005,hafemann2017offline,blanco2014performance}. In the digital age, handwritten signature verification remains a critical means of identity verification\cite{ozyurt2024offline,sanmorino2012survey,radhika2009pattern,diaz2019perspective}. However, the accuracy of signature verification faces significant challenges, primarily due to high intra-user variability and the risk of sophisticated forgery attacks \cite{Luiz2019DBLP,singer2016study,hafemann2017offline,blanco2014performance,sanmorino2012survey,radhika2009pattern,diaz2019perspective}.  

Signature verification is typically categorized into offline and online approaches based on the processing method\cite{hafemann2017offline,sanmorino2012survey,diaz2019perspective}. Offline signature verification primarily relies on analyzing static images, which involves global and local feature extraction and matching \cite{hafemann2017offline,Avola2021,pham2015offline,ruiz2008offline,huang2023multiscale}. However, offline methods are constrained by image quality and are highly sensitive to variations in writing styles and image capture conditions, leading to weaker anti-spoofing capabilities \cite{Epishkina2020,hafemann2017offline,pham2015offline}. In contrast, online signature verification (OSV) has emerged as the mainstream approach due to its capability to capture dynamic features such as pressure, velocity, and acceleration\cite{chandra2021novel,diaz2025neural,yahyatabar2017online,sekhar2020deepfuseosv}. These dynamic features provide richer information and exhibit significantly stronger anti-spoofing capabilities than offline methods \cite{Okawa2021,diaz2025neural,yahyatabar2017online,yang2024dynamic}.  

In recent years, deep learning techniques, particularly those leveraging deep metric learning, have shown significant potential in dynamic signature verification \cite{Rantzsch2016, Lai2017, Xie2022}. By optimizing the feature space to minimize intra-author distances and maximize inter-author distances, deep learning methods effectively reduce intra-user variability and enhance verification accuracy. However, existing studies primarily rely on Convolutional Neural Networks (CNNs) \cite{Lecun1998CNN} or Recurrent Neural Networks (RNNs) \cite{Jeffrey1990RNN}. Although these methods have made progress in feature extraction, they struggle to model temporal dependencies and often fail to capture the inherent dynamic patterns of signatures \cite{Shariatmadari2019, Wei2019, Lai2018, Li2019, Tolosana2017}.

\IEEEpubidadjcol

A signature can be effectively modeled as a sequence of strokes forming a graph, where each stroke is represented by multiple nodes, while the relationships between strokes are captured by edges \cite{Badie2024Offline,Roy2023offline}. The Graph Neural Networks (GNNs) are well-suited for capturing the complex relationships between nodes in graph-structured data, providing a promising approach for dynamic signature verification \cite{Badie2024Offline, Roy2023offline,maergner2018offline}. Inspired by the success of GNNs, we propose the Temporal-Spatial Graph Attention Transformer (TS-GATR), a novel framework that synergistically integrates graph-based spatial reasoning and sequential temporal modeling. Building upon the Graph Attention Layer (GAL) \cite{Xu2019MultiGraph}, which enhances Graph Attention Networks (GAT) \cite{Velickovic2018GAT} with transformer-style self-attention, TS-GATR introduces a Dual-Graph Attention Transformer (DGATR) module. This module constructs two complementary graph representations: a k-step adjacency graph to capture local stroke trajectories, and a k-nearest neighbor (k-NN) graph to model global geometric relationships. 

To capture temporal dependencies inherent in signature sequences, TS-GATR further incorporates a Gated Recurrent Unit (GRU) \cite{Cho2014GRU}. This design overcomes the limitation of traditional methods that struggle with long-range sequential dependencies, particularly in signatures with variable writing speeds or intermittent pauses. The temporal and spatial features are then fused to enhance the model’s ability to capture the underlying dynamics of the signature, leading to significant improvements in verification performance. By integrating Dynamic Time Warping (DTW)\cite{Müller2007DTW} as a distance measure, the TS-GATR method further enhances the accuracy and robustness of dynamic signature verification, effectively addressing the challenges posed by intra-user variability and forgery attacks.

The main contributions of this study are as follows:  

\begin{itemize}  
    \item We propose the Temporal-Spatial Graph Attention Transformer, a dual-branch architecture integrating the Dual-Graph Attention Transformer (DGATR) for spatial modeling and the Gated Recurrent Unit for temporal dynamics. This design jointly captures temporal-spatial dependencies in signatures, overcoming the limitations of prior single-modality approaches (e.g., CNNs or RNNs).  
    \item The DGATR module employs a dual-graph fusion strategy, combining \(k\)-step adjacency graphs and \(k\)-nearest neighbor adjacency graphs, to simultaneously model local stroke features and global spatial relationships, thereby enhancing feature representation.  
    \item Through extensive experiments on benchmark signature verification datasets, we demonstrate the superiority of the proposed TS-GATR method over existing state-of-the-art techniques, including DTW-based approaches and other deep learning models.  
\end{itemize}  

The rest of the paper is structured as follows. Section \ref{sec:related} surveys existing methodologies in dynamic signature verification, spanning classical parameter-based approaches, function-based temporal analysis, and contemporary deep learning frameworks. Section \ref{sec:method} introduces the Temporal-Spatial Graph Attention Transformer, elaborating on its dual-graph architecture for spatial modeling, temporal modeling with GRU, and hybrid loss functions. Section \ref{sec:experiment} evaluates the framework through comparative experiments on MSDS and DeepSignDB datasets, including ablation studies, parameter sensitivity analyses, and benchmark comparisons. Section \ref{sec:conclusion} concludes with insights into the model’s applicability and suggests future research directions.
\section{Related Works}
\label{sec:related}

Handwritten signature verification, as a critical technology in identity authentication and anti-counterfeiting, primarily employs two methodological approaches: parameter-based methods and function-based methods\cite{impedovo2012handwritten,parodi2013orthogonal,jia2019two,Jiang2022DsDTW}. Parameter-based methods verify signatures by extracting static geometric features (e.g., stroke length, angles)\cite{SaeBae2014, Guru2017}. While computationally efficient, they fail to effectively capture dynamic temporal variations, thereby limiting their performance in real-world scenarios. In contrast, function-based methods model signatures as time series that directly encode dynamic features, including positional trajectories, velocity, and acceleration \cite{okado2024dynamic,impedovo2012handwritten,parodi2013orthogonal,Jain2002,fermanian2021embedding}. These methods significantly improve the ability to handle temporal variations in signatures and have become the dominant paradigm in the field\cite{impedovo2012handwritten,Jain2002,fermanian2021embedding}.

Function-based methods typically employ machine learning techniques coupled with hybrid strategies that integrate both global and dynamic temporal features\cite{Chandra2022LWL,Mohammad2018Descriptor,Chandra2020OSVFuseNet,samatha2023deep}. For instance, Chandra et al. \cite{Chandra2020OSVFuseNet} developed a framework that integrates Hidden Markov Models (HMM)\cite{rabiner1986hhm,devijver1982pattern} with DTW. This approach utilizes global features to characterize signatures while resolving temporal misalignment via sequence alignment. The framework further incorporates Sequential Forward Feature Selection (SFFS)\cite{devijver1982pattern} to enhance cross-device compatibility. To address individual signature style variations, Sarvabhatla et al. \cite{Mohammad2018Descriptor} proposed an author-adaptive framework that clusters signature features using K-Means and selects optimal classifier-feature combinations through Equal Error Rate (EER) optimization, demonstrating significant accuracy improvements on the MCYT dataset\cite{ortega2003mcyt}. Furthermore, Locally Weighted Learning (LWL) \cite{Chandra2022LWL} was implemented to build local models capturing dynamic features such as pen pressure and acceleration, attaining a False Acceptance Rate (FAR) of 1.18\% and a False Rejection Rate (FRR) of 0.02\% on the SVC2004 dataset\cite{yeung2004svc2004}. Nevertheless, these approaches exhibit strong dependency on manual feature engineering, which constrains their generalization capability for complex temporal-spatial variations.

Deep learning has transformed signature verification by enabling end-to-end feature learning. Early research efforts primarily concentrated on integrating handcrafted features with deep neural network representations\cite{Chandra2020OSVFuseNet, Mohammad2018Descriptor}. As a representative example, OSVFuseNet \cite{Chandra2020OSVFuseNet} integrates Depthwise Separable Convolutional Neural Networks (DS-CNNs) with hand-engineered features, achieving high-precision classification under few-shot learning conditions. Unsupervised approaches, particularly sparse autoencoders \cite{Mohammad2018Descriptor}, learn invariant descriptors from unlabeled dual-channel (time-pressure) images and subsequently detect forged signatures through one-class classification. Recurrent neural architectures, such as Gated Recurrent Units (GRUs), reduce intra-class variations by jointly optimizing triplet loss and center loss, while simultaneously improving scale and rotation invariance through Length-Normalized Path Signature (LNPS) descriptors \cite{Lai2017LNPS}.

Dynamic Time Warping \cite{Müller2007DTW} is recognized as a fundamental technique for time series similarity measurement \cite{Plamondon1989}, serving as a cornerstone for aligning variable-length temporal sequences. Although DTW effectively addresses temporal distortions and sequence length discrepancies, its efficacy is significantly compromised in the presence of noise, outliers, and inherent signature variability \cite{Jiang2022DsDTW}. To augment its discriminative power, various enhanced methodologies have been developed, such as constrained alignment mechanisms \cite{Kar2017} and cost matrix optimization strategies \cite{Sharma2016}. Parziale et al. \cite{Parziale2019} proposed Stability-Modulated DTW (SM-DTW), which integrates stability regions into the DTW distance metric. Concurrently, Xia et al. \cite{Xia2018} devised a curve-constrained DTW variant specifically tailored for signature verification tasks. These enhanced approaches preserve DTW's core alignment advantages while effectively reducing sensitivity to noise and intra-class variability.

In the field of deep learning, Siamese networks exhibit strong capability in online signature verification through automated learning of similarity metrics between signature pairs. Wu et al. (2019) introduced two key innovations: the Pre-aligned Siamese Network (PSN) \cite{Wu2019PSN} and Deep Dynamic Time Warping (DDTW) \cite{Wu2019DDTW}. The PSN architecture utilizes DTW as a preprocessing module for temporal sequence alignment, while DDTW incorporates DTW as a differentiable network component to enable joint optimization of temporal alignment and feature representation learning. This dual mechanism effectively minimizes intra-class variance while maximizing inter-class separability. Building upon this, the Sig2Vec model \cite{Lai2022SynSig2Vec} leverages multi-head attention mechanisms to capture hierarchical temporal patterns in signature sequences, attaining state-of-the-art results on the DeepSignDB benchmark. The Time-Aligned Recurrent Neural Network (TA-RNN) \cite{DeepSignDB} incorporates DTW for optimal alignment of 23 temporal features derived from signature data, thereby improving the robustness of online verification systems. Nevertheless, the integration of conventional DTW with deep neural architectures frequently encounters optimization challenges due to their incompatible gradient computation mechanisms. To overcome this limitation, soft-DTW \cite{Jiang2022DsDTW}—a differentiable DTW variant—has been developed to facilitate end-to-end training of deep architectures. This approach retains DTW's temporal alignment capabilities while significantly boosting discriminative performance.

Despite these progressions, the accurate modeling of temporal-spatial dependencies in signatures continues to pose significant challenges\cite{DeepSignDB,Wu2019DDTW,Lai2022SynSig2Vec,Jiang2022DsDTW}. Current approaches predominantly depend on either conventional feature engineering or deep learning architectures that exhibit limited capability in handling complex sequential patterns and irregular temporal variations. To overcome these limitations, we propose a novel framework that synergistically integrates DGATR and GRU. This architecture enables joint learning of temporal-spatial features, thereby more effectively capturing both dynamic behavioral patterns and structural dependencies inherent in signature data. The proposed TS-GATR advances verification performance through synergistic integration of graph-based representation learning and sequential pattern modeling, effectively addressing the limitations inherent in both conventional and deep learning methodologies.

\section{Method}
\label{sec:method}

\begin{figure*}[t]
    \centering
    \includegraphics[width=1.0\linewidth]{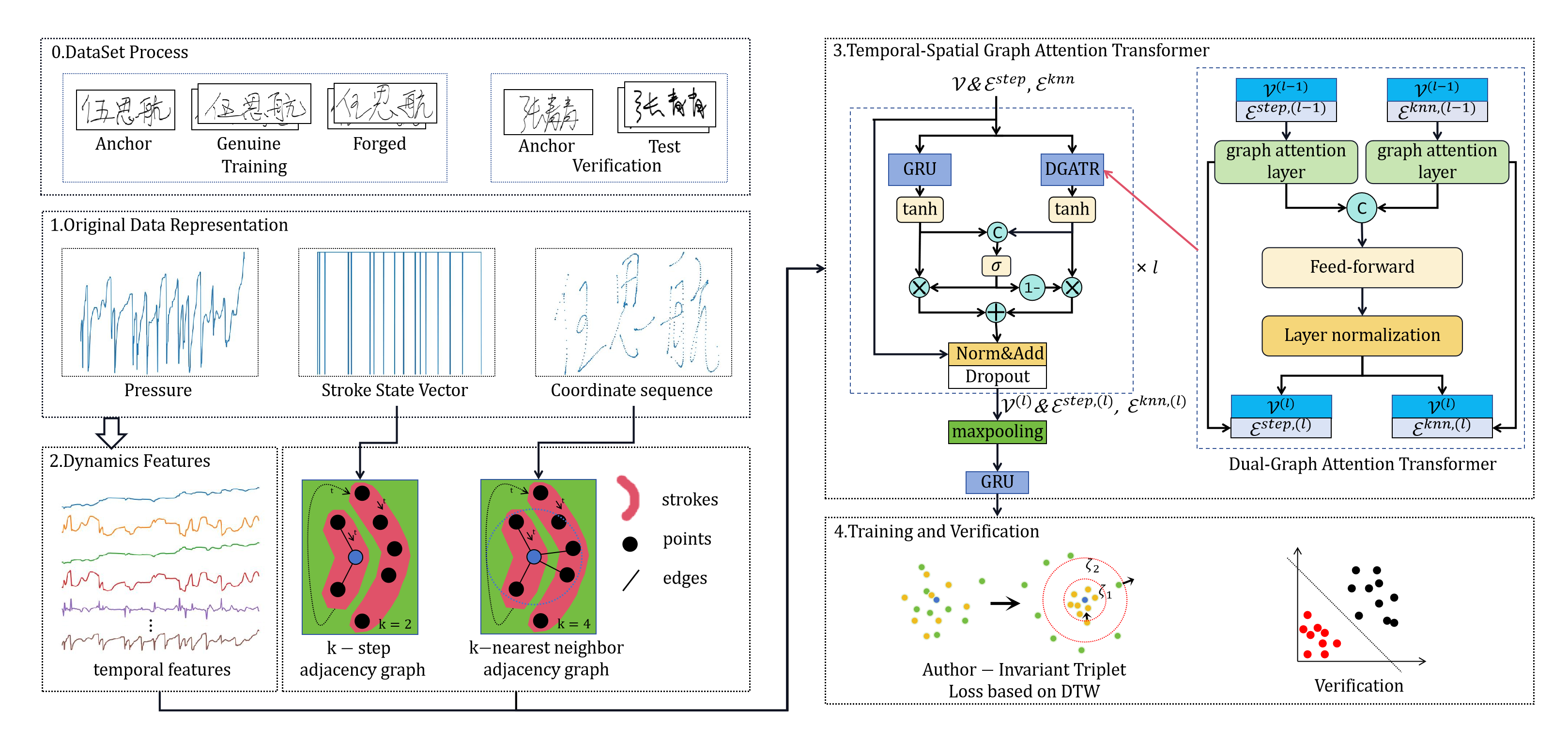}
    \caption{The overall pipeline of the proposed TS-GATR model combines DGATR with GRU to fuse temporal and spatial features.}
    \label{fig:pipeline}
\end{figure*}

Signature verification faces significant challenges in discriminative feature extraction due to writing style variations and subtle genuine-forged discrepancies. As shown in Fig.\ref{fig:pipeline}, we propose the Temporal-Spatial Graph Attention Transformer, an end-to-end framework that synergistically preserves spatial and temporal signature characteristics. This framework processes four signature modalities (pen coordinates, pressure, timestamps, stroke states) through two parallel representation schemes: (1) dual graph representations where k-step graphs model local stroke dynamics from stroke states, while k-NN graphs capture global geometric relationships through spatial proximity of pen coordinates; and (2) 16-dimensional dynamic features that are mathematically derived from all four raw modalities to comprehensively characterize writing kinematics.

The architecture employs parallel processing branches - the DGATR extracts multi-scale spatial patterns through fused k-step/k-NN graph features, while the GRU models temporal writing evolution. A gated fusion mechanism \cite{arevalo2017GMU} dynamically balances temporal-spatial feature contributions. During training, DTW aligns feature sequences and triplet loss optimizes genuine/forged separability. Verification follows DsDTW's threshold-based protocol \cite{Jiang2022DsDTW}. This unified approach enables robust verification against both natural variability and skilled forgeries.


\subsection{Symbol Description}

An online signature sample is represented as a multivariate time series \( \mathcal{S} = [\mathbf{C}, \mathbf{p}, \mathbf{t}, \mathbf{F}] \in \mathbb{R}^{L \times 5} \), where \( L \) denotes the total number of sampled points. The spatial trajectory is encoded in the coordinate matrix \( \mathbf{C} = [(x_1, y_1), \dots, (x_L, y_L)]^\top \in \mathbb{R}^{L \times 2} \), with \( (x_i, y_i) \) representing normalized pen coordinates. Temporal dynamics are captured by the pressure vector \( \mathbf{p} \in \mathbb{R}^L \) and timestamp vector \( \mathbf{t} \in \mathbb{R}^L \), where \( t_i \) denotes the cumulative time from the first sampled point. The vector \( F \in \mathbb{R}^{L} \), denoted as \( F = [f_1, f_2, \dots, f_L]^\top \), encodes sequential stroke dynamics for a sampled trajectory. Each component \( f_i \in \{0, 1, 2\} \) represents a discrete state indicating the phase of pen movement: initiation (\( f_i = 1 \)) marks the beginning of a stroke, continuation (\( f_i = 0 \)) denotes sustained contact during a stroke, and termination (\( f_i = 2 \)) signals the end of a stroke. This tri-state encoding explicitly segments the signature into contiguous strokes by identifying critical boundaries between successive pen-down and pen-up actions, while preserving intra-stroke continuity for modeling temporal coherence within individual strokes.

\subsection{Graph Attention Layer} \label{GAL}
\begin{figure}[t]
\centering
\includegraphics[width=0.8\linewidth]{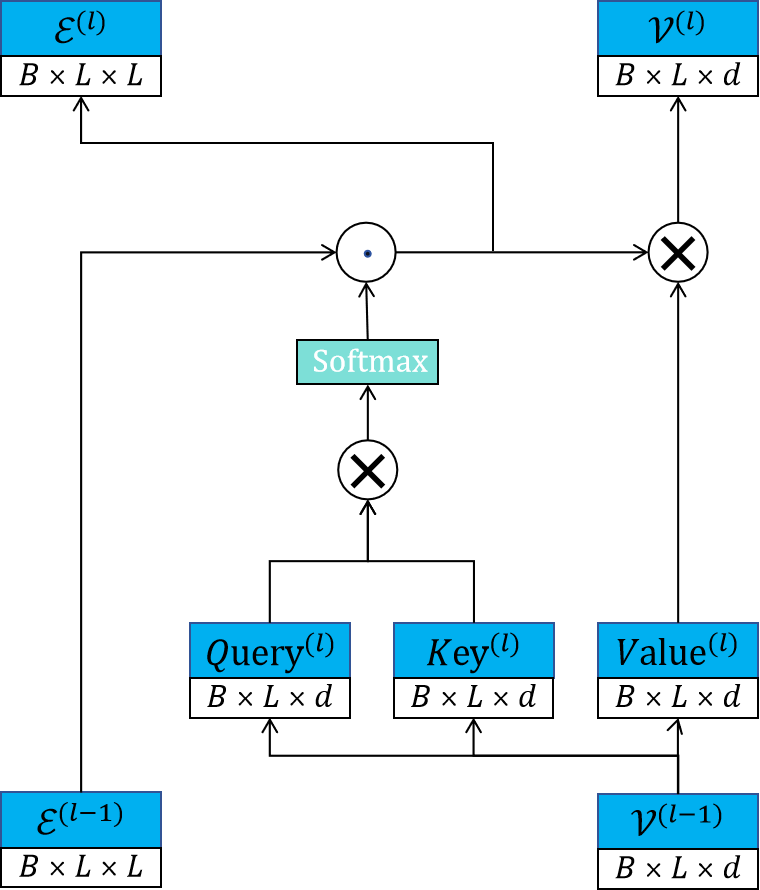}
\caption{Overview of the Graph Attention Layer (GAL). The architecture integrates static topological constraints with dynamic attention propagation. The design preserves graph topology while enabling adaptive relationship modeling. B, L and d denote batch size, node count, and feature dimension, respectively.}
\label{fig:gal}
\end{figure}

To model the non-Euclidean structural dependencies in signature graphs, we propose a Graph Attention Layer (GAL) that integrates static topological constraints with dynamic attention propagation. As shown in Fig.\ref{fig:gal}, the architecture processes graph-structured inputs \( \mathcal{G} = (\mathcal{V}, \mathcal{E}) \), and the node set \( \mathcal{V} = \{v_i\}_{i=1}^L \) consists of vector representations of sampled signature points, where each node \( v_i \in R^d \) is a \( d \)-dimensional feature vector, \( v_i \in \mathbb{R}^d \). The adjacency matrix \( \mathcal{E} \in \{0,1\}^{L \times L} \) explicitly encodes structural priors, with each element \( e_{ij} \in \{0,1\} \) indicating the presence (\( e_{ij} = 1 \)) or absence (\( e_{ij} = 0 \)) of a direct edge between nodes \( v_i \) and \( v_j \). 

The model operates on the principle of structure-constrained attention \cite{Xu2019MultiGraph}: the original adjacency matrix \( \mathcal{E} \) defines persistent neighborhoods \( \mathcal{N}(i) = \{j | e_{ij} = 1\} \), which remain invariant throughout network propagation. Within these static neighborhoods, layer-specific attention weights \( e_{ij}^{(l)} \) evolve to modulate interaction intensities between connected nodes. The feature update mechanism is formalized as:

\begin{align}
    \hat{e}_{ij}^{(l)} = \frac{W_Q^{(l)} v_i^{(l-1)} (W_K^{(l)} v_j^{(l-1)})^T}{\sqrt{d}} e_{ij}^{(l-1)},
\end{align}

\begin{align}
    e_{ij}^{(l)} = \frac{\exp(\hat{e}_{ij}^{(l)})}{\sum_{k \in \mathcal{N}(i)} \exp(\hat{e}_{ik}^{(l)})},
\end{align}

\begin{align}
    v_i^{(l)} = \sum_{j \in \mathcal{N}(i)} e_{ij}^{(l)} W_V^{(l)} v_j^{(l-1)},
\end{align}
where \( e_{ij}^{(l)} \) represents the attention weight along edge \( (i,j) \) at layer \( l \), initialized as \( e_{ij}^{(0)} = e_{ij} \). The transformation matrices \( W_Q^{(l)}, W_K^{(l)}, W_V^{(l)} \in \mathbb{R}^{d \times d} \) denote the learnable projection parameters for query, key, and value representations, respectively. The softmax function is strictly applied over the fixed neighborhood \( \mathcal{N}(i) \), ensuring that structural dependencies are preserved while allowing the model to learn hierarchical attention patterns. 

The architecture achieves two critical objectives: 1) Structural preservation through persistent neighborhoods prevents over-smoothing and maintains interpretable graph topology; 2) Contextual adaptation via layer-wise attention weights enables dynamic relationship modeling. This dual mechanism is particularly effective in signature verification, where stable structural patterns coexist with subtle, context-dependent variations. The fixed neighborhood constraint not only aligns with the intrinsic sparsity of signature graphs but also provides inductive biases for robust feature learning.

\subsection{Graph Structure Representation}

To establish a hierarchical representation of online signatures, we formulate the dynamic signature as a graph \( \mathcal{G} = (\mathcal{V}, \mathcal{E}) \). We propose two complementary adjacency graph construction paradigms that explicitly model intra-stroke continuity and inter-stroke correlations using distinct topological constraints.

\textbf{k-Step Adjacency Graph \cite{Xu2019MultiGraph}:}  
This graph structure captures sequential dependencies within individual strokes by enforcing temporal proximity constraints. Let \( \mathcal{V}_m \subset \mathcal{V} \) denote the node subset belonging to the \( m \)-th stroke, determined by the label vector \( F \). The adjacency matrix \( \mathcal{E}^{\text{step}} \) is constructed as:

\begin{align}
    e^{\text{step}}_{ij} = \begin{cases} 
1 & \text{if } \exists m \text{ s.t. } v_i, v_j \in \mathcal{V}_m \text{ and } 0 \leq |i-j| \leq k ,\\
0 & \text{otherwise},
\end{cases}
\end{align}
where \( k \) controls the maximum temporal distance between connected nodes. As illustrated in Fig.\ref{fig:pipeline}, the resultant graph forms a directed chain structure within each stroke, preserving the original temporal order of sampling points while eliminating spurious connections across stroke boundaries. This design effectively encodes local kinematic patterns such as velocity transitions and acceleration profiles within individual pen movements.

\textbf{k-Nearest Neighbor (k-NN) Adjacency Graph:}  
To model global spatial relationships that transcend stroke segmentation, we construct an undirected graph based on Euclidean spatial proximity. For each node \( v_i \) with coordinates \( (x_i, y_i) \), compute pairwise distances \( d_{ij} = \| (x_i, y_i) - (x_j, y_j) \|_2 \) and establish edges to its \( k \) nearest neighbors:

\begin{align}
    e^{\text{knn}}_{ij} = \begin{cases} 
1 & \text{if } v_j \in \text{Top}_k(\{d_{i1},...,d_{iL}\}) ,\\
0 & \text{otherwise} ,
    \end{cases}
\end{align}
where \( \mathcal{E}^{\text{knn}} \) dynamically adapts to spatial density variations by establishing connections based on geometric proximity. As illustrated in Fig.\ref{fig:pipeline}, this graph topology captures critical geometric features — including stroke intersections, spatial clustering patterns, and global contour geometries — through connections that transcend temporal stroke segmentation. The adaptive nature of k-NN connectivity ensures structural consistency under non-uniform sampling: in sparsely sampled regions, nodes automatically form extended connections to preserve topological integrity, while maintaining localized interactions in dense areas. Crucially, this framework enables cross-stroke relational reasoning by linking spatially adjacent nodes from temporally disjoint strokes, thereby modeling latent spatial dependencies essential for holistic signature analysis.

The dual-graph architecture provides complementary perspectives for signature representation:  
1) The k-step graph preserves temporal locality and stroke isolation through hard structural constraints, modeling fine-grained motion dynamics.  
2) The k-NN graph introduces static spatial proximity constraints that reveal latent geometric correlations across strokes, essential for forgery verification requiring global shape analysis.  

\subsection{Dual-Graph Attention Transformer}  \label{DGATR}
To integrate complementary information from both graph modalities, we design a fusion framework \cite{Xu2019MultiGraph} that jointly leverages local and global spatial relationships. Separate GAL modules process each graph structure independently:  

\begin{align}
    \mathcal{V}_{\mathcal{E}}^{(l)},\mathcal{E}^{(l)} = GAL(\mathcal{V}^{(l-1)}, \mathcal{E}^{(l-1)}),
\end{align}
where \( GAL \) denotes the graph attention layer defined in \ref{GAL}.

The fused representation is generated by concatenating the outputs of both branches, followed by a Feed-Forward Network (FFN) implemented as a learnable linear projection \(W_c \in \mathbb{R}^{2d \times d}\) and a nonlinear activation \(\sigma(\cdot)\):  
\[
\mathcal{V}^{(l)} = \sigma\left( W_c \left[ \mathcal{V}_{\mathcal{E}^{\text{knn}}}^{(l)} \Vert \mathcal{V}_{\mathcal{E}^{\text{step}}}^{(l)} \right] \right).  
\]  
Here, \( \Vert \) denotes feature concatenation, \(W_c\) constitutes the core component of the FFN, mapping concatenated features to a unified subspace. Layer normalization \cite{ba2016layernormalization} is subsequently applied to stabilize training dynamics. This complementary fusion balances stroke-localized dynamics from \( \mathcal{E}^{\text{step}} \) and cross-stroke geometric patterns from \( \mathcal{E}^{\text{knn}} \), enhancing discriminative power for subtle signature variations. 

\subsection{Temporal-spatial Collaborative Representation Learning}

Online signatures inherently exhibit coupled temporal-spatial dynamics where stroke geometry and writing dynamics evolve interdependently. To jointly model these interdependent patterns, we propose a Temporal-Spatial Graph Attention Transformer featuring a structure-dynamics dual-branch architecture, with two key modules: 1) Parallel pathways for spatial structure encoding and temporal dynamics modeling, and 2) A gated fusion mechanism \cite{arevalo2017GMU} for cross-modal feature complementarity. This dual-branch architecture jointly captures structural stability (e.g., stroke connectivity patterns) and dynamic continuity (e.g., acceleration profiles) in handwriting biometrics.


The model employs independent branches to process spatial and temporal features. Spatial Structure Branch inherits the multi-scale topology modeling capabilities of the DGATR module (\ref{DGATR}). It simultaneously extracts intra-stroke sequential patterns and inter-stroke geometric relationships through dual-graph attention. For layer \( l \), the processing is formalized as:
\begin{align}
    \mathcal{V}^{(l)}, & \mathcal{E}^{\text{step},(l)}, \mathcal{E}^{\text{knn},(l)} = \notag \\
    & DGATR( \mathcal{V}^{(l-1)}, \mathcal{E}^{\text{step},(l-1)},  \mathcal{E}^{\text{knn},(l-1)}),
\end{align}
Here, \(  \mathcal{E}^{\text{step},(l-1)} \) constrains sequential dependencies within strokes, while \( \mathcal{E}^{\text{knn},(l-1)} \) models cross-stroke spatial proximity via k-nearest neighbors. 

Temporal Dynamics Branch utilizes Gated Recurrent Unit (GRU) to capture continuity in writing motion. For layer \( l \), the state update emphasizes dynamic feature extraction:
\begin{align}
    D^{(l)} = GRU(\mathcal{V}^{(l-1)}),
\end{align}
where \( D^{(l)} \in \mathbb{R}^{L \times d} \) represents the GRU’s hidden states. Through coordinated operation of reset gates and update gates, this branch adaptively track temporal context evolution, focusing on rhythm variations.


To achieve effective alignment and complementary integration of temporal-spatial features, we adapt a gated fusion mechanism \cite{arevalo2017GMU} with residual enhancement. The fusion process operates through two sequential operations:

\begin{equation}
    Z^{(l)} = \sigma\left( W_g [\mathcal{V}^{(l)} \Vert D^{(l)}] \right),
\end{equation}

\begin{equation}
    \mathcal{V}^{(l)} \gets Z^{(l)} \odot \mathcal{V}^{(l)} + (1 - Z^{(l)}) \odot D^{(l)} + \mathcal{V}^{(l-1)},
\end{equation}
Here, \( W_g \in \mathbb{R}^{2d \times d} \) are learnable parameters that project the concatenated temporal-spatial features into a unified representation space. The sigmoid activation $\sigma(\cdot)$ generates element-wise interpolation weights $Z^{(l)} \in [0,1]^{d}$ for feature blending, with $\odot$ denotes element-wise multiplication. To further enhance the expressive power of the fused features and mitigate gradient vanishing issues in deep networks, we introduce residual connections during the fusion process\cite{he2015res}.

As depicted in Fig.\ref{fig:pipeline}, TS-GATR iteratively refines temporal-spatial representations through alternating feature abstraction and modality fusion.This dual-branch design enables comprehensive representation of both microscopic motion dynamics and macroscopic spatial patterns, effectively addressing the challenge of skilled forgeries where attackers often replicate geometric features but fail to mimic authentic kinematic and rhythmic signatures.

\subsection{Time-Functions Extraction}

For signature data \( \mathcal{S} = [\mathbf{C}, \mathbf{p}, \mathbf{t}, \mathbf{F}] \), to capture the temporal characteristics of the writing process, we extract 12 time functions to derive writing features and centralize them\cite{Jiang2022DsDTW,DeepSignDB}:

\par - pen pressure: \( p \)
\par - first-order derivative of x,  y-coordinate: \( \dot{x}_n \text{,  } \dot{y}_n \)
\par - velocity magnitude: \( v_n =\sqrt{(\dot{x}_n^2+\dot{y}_n^2)} \)
\par - path-tangent angle: \( \theta_n = arctan(\dot{y}_n / \dot{x}_n) \)
\par - \( cos(\theta_n)  \text{,  } \sin(\theta_n) \)
\par - first-order derivative of \( v_n \text{,  } \theta_n \): \( \dot{v}_n \text{,  } \dot{\theta}_n \)
\par - log curvature radius: \( \rho_n = log(v_n / \theta_n ) \)
\par - centripetal acceleration: \( a_c \)
\par - total acceleration magnitude: \( a_n \)


\subsection{Dynamic Time Warping}

Dynamic Time Warping \cite{Müller2007DTW} is an algorithm used to measure the similarity between two time series. Its core idea is to use a nonlinear alignment method that allows for different stretching along the time axis, thus providing a precise similarity measure between sequences. In the context of signature data, DTW can handle time-scale differences due to variations in writing speed, pressure, and other factors, allowing for more accurate similarity measurement between samples.

Given two time series \( X = [ x_1, x_2, \dots, x_{L_1} ] \in \mathbb{R}^{L_1 \times d} \) and \( Y = [ y_1, y_2, \dots, y_{L_2} ] \in \mathbb{R}^{L_2 \times d} \), where each time series consists of \( L \) data points, with each data point being a \( d \)-dimensional feature vector, the goal of DTW is to minimize the cumulative distance between the two sequences by introducing nonlinear matching, defined as:
\begin{align}
    \text{dtw}(X, Y) = \min_{\pi \in \Pi} \sum_{t=1}^{L_1} \| x_t - y_{\pi(t)} \|^2,
\end{align}

subject to:

\begin{align}
    \pi(1) = 1, \ \pi(L_1) = L_2, \ \pi(t+1) \geq \pi(t),
\end{align}
where \( \Pi \) is the set of all possible nonlinear matching paths, and \( \pi(t) \) denotes the index of the element in sequence \( Y \) that matches with the \( t \)-th point in sequence \( X \).

To solve this problem, DTW is typically optimized via dynamic programming. The alignment cost between two time series is recursively computed as follows:  
\begin{align}
\text{dtw}(X_{1:t}, &Y_{1:s}) = \|x_t - y_s\|^2 + \min\{\text{dtw}(X_{1:t-1}, Y_{1:s}),  \notag \\
&\text{dtw}(X_{1:t}, Y_{1:s-1}), \text{dtw}(X_{1:t-1}, Y_{1:s-1})\}
\end{align} 
where \( \text{dtw}(X_{1:t}, Y_{1:s}) \) represents the minimum accumulated alignment cost up to time indices \( t \) and \( s \) of the input sequences.  

The advantage of DTW lies in its ability to capture nonlinear deformations between time series, particularly in signature recognition tasks. Even when there are changes in the time, speed, or rhythm of the strokes, DTW can still accurately measure their similarity. Thus, in this study, we use DTW to measure the feature distances between signature samples, providing an effective metric for the design of the loss function.

To enhance DTW's computational efficiency while preserving its alignment accuracy, we introduce a hybrid module combining pooling and GRU layers at the final stage of our backbone network. The max-pooling layer performs stride-2 temporal subsampling, reducing sequence length by 50\%, while preserving full feature dimensionality.  This is followed by a GRU layer that selectively preserves critical temporal dynamics through its gating mechanism, maintaining essential stroke rhythm and timing patterns for skilled forgery detection. 

\subsection{Author-Invariant Triplet Loss}

To enhance the generalization capability of handwritten signature authenticity verification models, this study proposes a joint loss function that integrates DTW with a dual-constraint mechanism. This method effectively distinguishes genuine signatures from two types of forgeries (random Forgery or Skilled Forgery) by constructing a unified feature space across authors. For each author \( p \), based on their sub-dataset \( D_p = \{\{\mathcal{S}^{a_i}_p\}_{i=1}^{n_a}, \{\mathcal{S}^{g_j}_p\}_{j=1}^{n_g}, \{\mathcal{S}^{f_k}_p\}_{k=1}^{n_f}\} \), the model simultaneously optimizes the following two objectives during training:

Relative Margin Loss: This loss function enforces discriminative feature learning by constraining the distance difference between genuine and forged signatures relative to the anchor. Given the anchor signature representation \( H^a_p \), a genuine sample \( H^g_p \), and a forged sample \( H^f_p \), the loss is defined as:  
\begin{align}
\mathcal{L}_{m,p}^{(a,g,f)} = \text{ReLU}\left( \text{dtw}(H^a_p, H^g_p) - \text{dtw}(H^a_p, H^f_p) + \gamma_1 \right) ,
\end{align}
where \( \gamma_1 \) is a predefined margin parameter. This constraint ensures that the DTW distance between forged signatures and the anchor is at least \( \gamma_1 \) greater than that of genuine signatures, thereby establishing a secure decision boundary in the feature space.

Pairwise Threshold Loss: To eliminate threshold bias across authors, a global similarity constraint is designed. For the anchor representation \( H^a_k \) and a test sample \( H^s_k \) (either genuine or forged), the loss is computed as:  
\begin{align}
\mathcal{L}^{(a,s)}_{\text{th,p}} = \text{ReLU}\left( \xi - l_s \cdot (\gamma_2 - \text{dtw}(H^a_p, H^s_p)) \right),
\end{align}  
where \( l_s \in \{1, -1\} \) is a class indicator (1 for genuine, -1 for forged), and \( \gamma_2 \) represents a unified similarity threshold across authors. This function regulates the global similarity baseline through \( \gamma_2 \), with \( \xi \) acting as a relaxation factor to accommodate intra-author variations.

A linear weighting strategy is adopted to synergize the advantages of the two constraints. The total objective function is formulated as:  
\begin{align}
\mathcal{L}_{\text{p}} = \alpha \cdot \sum_{i,j,k} \mathcal{L}_{m,p}^{(a_i,g_j,f_k)} + \beta \cdot \left( \sum_{i,j} \mathcal{L}_{\text{th,p}}^{(a_i,g_j)} + \sum_{i,k} \mathcal{L}_{\text{th,p}}^{(a_i,f_k)} \right),
\end{align} 
where \( \alpha \) and \( \beta \) are balancing hyperparameters. This design achieves dual-mechanism synergy: (1) \( \mathcal{L}_{m,p}^{(a,g,f)} \) enhances intra-author discriminability between positive and negative samples, while (2) \( \mathcal{L}^{(a,s)}_{\text{th,p}} \) establishes a cross-author unified decision standard to suppress threshold drift caused by stylistic variations.  

\section{Experiments}
\label{sec:experiment}

\subsection{Dataset Overview}

In this study, we propose the TS-GATR model, which integrates GNN and GRU to effectively capture both temporal dynamics and spatial structural features in signature data. To validate the efficacy of this approach, we conducted extensive experiments on two distinct datasets, i.e., MSDS\cite{zhang2022msds} and DeepSignDB\cite{DeepSignDB}.

\begin{itemize} 
\item \textbf{MSDS Dataset}: This dataset comprises handwritten signature samples from 402 users, with each user providing 20 genuine and 20 forged signatures. Data collection spanned two sessions, with a minimum interval of 21 days between them. The dataset is divided into two subsets: MSDS-ChS (Chinese signatures) and MSDS-TDS (Token Digit Strings). Each signature includes both time-series data and static image data. \item \textbf{DeepSignDB Dataset}: This dataset contains 69,972 handwritten signatures from 1,526 authors, distributed across five subsets: MCYT, BiosecurID, Biosecure DS2, e-BioSign DS1, and e-BioSign DS2. Unlike the MSDS dataset, DeepSignDB includes signatures written with both stylus and finger inputs. Additionally, it provides standardized evaluation protocols to ensure consistent experimental conditions and result comparisons\cite{DeepSignDB,Jiang2022DsDTW}. 
\end{itemize}

\subsection{Experimental Setup}

To assess the performance of our proposed TS-GATR model, we established the following experimental setup:

\begin{itemize}
\item \textbf{Data Preprocessing}: We performed denoising and normalization on the raw time-series data to ensure data quality. The time-series data encompass x and y coordinates, pressure values, and timestamps recorded during the writing process.

\item \textbf{Baseline Methods}: We compared the TS-GATR model with several existing baseline methods, including:
    \begin{itemize}
    \item \textbf{DTW \cite{Müller2007DTW}}: A technique used to measure similarity between time-series data, widely applied in signature verification.
    \item \textbf{Sig2Vec \cite{Lai2022SynSig2Vec}}: A signature verification model based on one-dimensional Convolutional Neural Networks (CNNs), which has demonstrated excellent performance on the DeepSignDB dataset.
    \item \textbf{TA-RNNs \cite{DeepSignDB}}: Time-Aligned Recurrent Neural Networks that combine Bidirectional Long Short-Term Memory (BiLSTM) networks for temporal alignment.
    \item \textbf{DsDTW \cite{Jiang2022DsDTW}}: A signature verification method based on soft Dynamic Time Warping (soft-DTW).
    \end{itemize}
\item \textbf{Evaluation Metrics}: We adopt Equal Error Rate \cite{Jiang2022DsDTW} as the primary evaluation metric, which represents the error rate at the point where the False Acceptance Rate equals the False Rejection Rate. Following the original papers of MSDS and DeepSignDB, we report EERs under both a global threshold and a local (user-specific) threshold, presenting results in the format of \(EER_g/EER_l\) for MSDS-ChS and MSDS-TDS \cite{zhang2022msds}. For DeepSignDB, only global threshold evaluations are reported. All results are expressed as percentages. Additionally, we evaluate the model against two types of impostor attacks: skilled and random forgeries. Skilled forgeries refer to expertly forged samples originally provided in the datasets, representing high-effort attacks, while random forgeries are generated by selecting genuine signatures from other users, simulating unauthorized access attempts by individuals unfamiliar with the target user’s signature style. This evaluation strategy ensures a comprehensive assessment of the model’s robustness against varying levels of adversarial threats.

\item \textbf{Training and Verification}: For the MSDS dataset, we divided 202 users into a training set and 200 users into a testing set \cite{zhang2022msds}. We then evaluated the models on the testing set to obtain EER. For the DeepSignDB dataset, we conducted experiments using the standard training and testing sets provided in \cite{DeepSignDB}.we designed two sets of parallel experiments. In the first set of experiments, we followed the setting of \cite{zhang2022msds, Jiang2022DsDTW} and selected the first $n$ samples from each user's real signature as signature templates. The second set of experiments aims to explore the impact of template selection on model performance. To this end, we conducted multiple experiments under different template selection conditions, specifically by randomly selecting different template configurations, repeating the experiment 20 times, and using the minimum value among the results as the final evaluation metric.
\end{itemize}


\subsection{Validation of Temporal-Spatial Modeling Components}

To evaluate the contributions of key components in the TS-GATR model, ablation studies were conducted on the MSDS-ChS and MSDS-TDS datasets. Three configurations were designed: (1) the full TS-GATR model integrating both GRU and DGATR modules, (2) a GRU-only variant with DGATR removed, and (3) a DGATR-only variant with GRU removed. The results are summarized in Table \ref{tab:ablation_study}.

\begin{table}[h]
    \centering
    \scriptsize 
    \renewcommand{\arraystretch}{1.2} 
    \setlength{\tabcolsep}{4pt} 
    \caption{Ablation study comparing the performance of TS-GATR with different configurations: (1) the full TS-GATR model with both GRU and DGATR, (2) GRU-only version, and (3) DGATR-only version.}
    \label{tab:ablation_study}
    \begin{tabular}{ccccccc}
        \toprule
        \multirow{2}{*}{Dataset} & \multirow{2}{*}{GRU} & \multirow{2}{*}{DGATR} & \multicolumn{2}{c}{Skilled Forgery} & \multicolumn{2}{c}{Random Forgery} \\
        \cmidrule(lr){4-5}  \cmidrule(lr){6-7}
        & & & 1vs1 & 4vs1 & 1vs1 & 4vs1 \\ \hline
        \multirow{3}{*}{MSDS-ChS}
    &\checkmark &\checkmark &\textbf{8.16/3.90}  & \textbf{6.22/2.77} & \textbf{2.28}/0.51           &\textbf{1.26/0.16} \\
    &\checkmark & &10.24/5.91 &7.52/4.51 &2.72/0.62 &1.84/0.43 \\
    & &\checkmark &9.36/4.59 &7.27/0.95 &2.43/\textbf{0.46} &1.47/0.33 \\ \hline

        \multirow{3}{*}{MSDS-TDS}
    &\checkmark &\checkmark &\textbf{5.26/2.42}&\textbf{4.23/1.78}  &1.79/0.34 &1.46/0.21 \\
    &\checkmark & &7.45/4.08 &6.61/2.87 &2.62/0.73 &2.05/0.46 \\
    & &\checkmark &5.84/1.97 &4.35/1.42 &\textbf{1.76/0.27} &\textbf{1.45/0.10} \\ \hline
        
    \end{tabular}
    
\end{table}

From the results in Table \ref{tab:ablation_study}, the full model achieves optimal performance on Skilled Forgery tasks (MSDS-ChS: 8.16/3.90 and 6.22/2.77; MSDS-TDS: 5.26/2.42 and 4.23/1.78), demonstrating the necessity of joint temporal-spatial modeling for capturing intricate forgery patterns. The GRU-only variant exhibits significant performance degradation in Skilled Forgery tasks (e.g., MSDS-ChS 1vs1 EER increases to 10.24/5.91), indicating that temporal modeling alone fails to capture spatial correlations in dynamic signatures. These results validate the complementary roles of GRU and DGATR: GRU captures temporal dynamics through gated mechanisms, while DGATR models spatial dependencies via graph structures, collectively enhancing discriminative power against multi-scale forgery features.

Notably, the DGATR-only configuration achieves competitive results in Skilled Forgery (e.g., MSDS-TDS 1vs1: 5.84/1.97) and excels in Random Forgery tasks (MSDS-TDS: 1.76/0.27 and 1.45/0.10). This aligns with the task characteristics: The Random Forgery sample set, generated by randomly combining signatures from different authors, exhibit global spatial discrepancies rather than fine-grained temporal inconsistencies. DGATR’s graph attention mechanism effectively identifies spatial heterogeneity without relying on temporal modeling. 

\subsection{Trade-off Between Pooling Strategies and Computational Efficiency}

To optimize DTW efficiency while preserving performance, we investigated pooling strategies and GRU configurations at the backbone network’s final stage. Four configurations were compared: (1) Average Pooling (AP) + GRU, (2) Max Pooling (MP) + GRU, (3) GRU-only, and (4) No Pooling/GRU. Results are shown in Table \ref{tab:pooling_study}.

\begin{table*}[h]
    \centering
    \renewcommand{\arraystretch}{1.2} 
    \setlength{\tabcolsep}{6pt} 
    \caption{Impact of pooling strategies and GRU configurations on Skilled/Random Forgery tasks. Four configurations were compared: (1) AP+ GRU, (2) MP + GRU, (3) GRU-only,  (4) No Pooling/GRU.}
    \label{tab:pooling_study}
    \begin{tabular}{cccccccc}
        \toprule
        \multirow{2}{*}{Dataset} & \multirow{2}{*}{MP} & \multirow{2}{*}{AP} & \multirow{2}{*}{GRU-POST} & \multicolumn{2}{c}{Skilled Forgery} & \multicolumn{2}{c}{Random Forgery} \\
        \cmidrule(lr){5-6}  \cmidrule(lr){7-8}
        & & & & 1vs1 & 4vs1 & 1vs1 & 4vs1 \\ \hline
        \multirow{4}{*}{MSDS-ChS}
        &\checkmark & &\checkmark &\textbf{8.16/3.90}  & 6.22/\textbf{2.77} & \textbf{2.28/0.51}           &\textbf{1.26/0.16} \\
        & &\checkmark &\checkmark &8.22/3.90 &\textbf{5.85}/2.80 &2.90/0.48 &1.70/0.39 \\ %
        & & &\checkmark &8.26/4.02 &5.93/2.85 &2.73/0.45 &2.31/0.64 \\ %
        & & & &11.91/8.09 &9.47/6.34 &\textbf{4.79/1.19} &\textbf{2.88/1.25} \\ \hline %

        \multirow{4}{*}{MSDS-TDS}
        &\checkmark & &\checkmark &\textbf{5.26/2.42}&4.23/1.78  &1.79/0.34         &1.46/0.21  \\
        & &\checkmark &\checkmark &5.55/2.26 &\textbf{4.12/1.57} &\textbf{1.66/0.26} &\textbf{1.46/0.15} \\ 
        & & &\checkmark &5.33/2.61&4.34/1.91  &1.86/0.41  &1.51/0.25 \\ 
        & & & &6.81/2.87 &4.94/2.07 &2.11/0.75  &1.75/0.37 \\ \hline
        
    \end{tabular}
\end{table*}

Table \ref{tab:pooling_study} demonstrates that the MP+GRU configuration achieves optimal Skilled Forgery performance (MSDS-ChS 1vs1: 8.16/3.90; MSDS-TDS 1vs1: 5.26/2.42). The AP + GRU variant performs comparably (e.g., MSDS-ChS 1vs1: 8.22/3.90), suggesting that both pooling methods enhance efficiency without significant information loss.This supports the hypothesis that pooling reduces computational redundancy while GRU compensates for potential feature degradation by retaining critical sequential patterns. 

\subsection{Parameter Sensitivity Analysis for Graph Structures}

On the MSDS dataset, we systematically investigated the parameter optimization mechanisms of k-NN and k-Step adjacency graphs and their impacts on signature verification performance. To control variable interference, k-Step parameters were fixed at k=2 when studying k-NN adjacency graphs, while optimal k-NN parameters (k=35 for Chinese signatures (MSDS-ChS) and k=20 for digit signatures (MSDS-TDS)) were set when analyzing k-Step graphs. For the MSDS-Chs dataset, we use a grid search strategy to optimize the hyperparameters, setting the neighborhood scale of k-NN to $k \in \{20, 30, 35, 40, 50\}$ and the time window depth (k-Step) to $k \in \{1, 2, 4, 6, 8\}$. For the MSDS-TDs dataset, the neighborhood scale is configured to $k \in \{10, 20, 25, 30, 40\}$ and the time window depth is also set to $k \in \{1, 2, 4, 6, 8\}$. The relevant experimental results are reported in Table \ref{tab:knnAdjacencyGraph} and Table \ref{tab:kstepAdjacencyGraph}, respectively.

\begin{table}[h]
    \centering
    \scriptsize 
    \renewcommand{\arraystretch}{1.2} 
    \setlength{\tabcolsep}{6pt} 
    \caption{Performance Analysis of \( k \)-Nearest Neighbor Adjacency Graphs in Signature Verification. The table shows Equal Error Rates (EER, \%) across neighborhood sizes (\( k = 10-50 \)) for Chinese (MSDS-ChS) and digit (MSDS-TDS) signatures under skilled and random forgery attacks.}
    \label{tab:knnAdjacencyGraph}
    \begin{tabular}{cccccc}
        \hline
        \multirow{2}{*}{Dataset} & \multirow{2}{*}{k} & \multicolumn{2}{c}{Skilled Forgery} & \multicolumn{2}{c}{Random Forgery} \\
        \cmidrule(lr){3-4}  \cmidrule(lr){5-6}
        & & 1vs1 & 4vs1 & 1vs1 & 4vs1 \\ \hline
        \multirow{5}{*}{MSDS-ChS}
& 20   & 8.39/4.15 & 6.25/3.06 & \textbf{2.18/0.36} & \textbf{1.28}/0.18 \\
& 30   & 8.47/4.48 &\textbf{6.20}/3.11 & 2.27/0.40 & 1.38/0.18 \\
& 35   &\textbf{8.16/3.90} &6.22/\textbf{2.77} &2.28/0.51 &\textbf{1.26/0.16} \\
& 40   & 8.62/4.46 & 6.38/3.19 & 2.54/0.38 & 1.56/0.16 \\  
& 50  & 8.51/4.10 & 6.45/2.89 & 2.67/0.45 & 1.58/0.23 \\ \hline
        \multirow{5}{*}{MSDS-TDS}
& 10   &5.53/2.70         & 4.38/1.85          &1.73/0.40         &1.50/0.25 \\  
& 20   &\textbf{5.26/2.42}&\textbf{4.23/1.78}  &1.79/0.34         &1.46/0.21 \\  
& 25   & 5.41/2.77        & 4.47/1.98          &1.85/0.37         & 1.54/0.23 \\
& 30   & 5.49/2.76        & 4.51/2.13          &1.74/0.38         &\textbf{1.38}/0.22 \\
& 40   & 5.32/2.53        & 4.51/2.00           &\textbf{1.71/0.30}&1.54/\textbf{0.18} \\ \hline
    \end{tabular}
\end{table}

\begin{table}[h]
    \centering
    \scriptsize 
    \renewcommand{\arraystretch}{1.2} 
    \setlength{\tabcolsep}{6pt} 
    \caption{Performance Analysis of \( k \)-Step Adjacency Graphs in Signature Verification. The table shows Equal Error Rates (EER, \%) across temporal depths (\( k = 1-8 \)) for Chinese (MSDS-ChS) and digit (MSDS-TDS) signatures under skilled and random forgery attacks.}
    \label{tab:kstepAdjacencyGraph}
    \begin{tabular}{cccccc}
        \hline
        \multirow{2}{*}{Dataset} & \multirow{2}{*}{k} & \multicolumn{2}{c}{Skilled Forgery} & \multicolumn{2}{c}{Random Forgery} \\
        \cmidrule(lr){3-4}  \cmidrule(lr){5-6}
        & & 1vs1 & 4vs1 & 1vs1 & 4vs1 \\ \hline
        \multirow{5}{*}{MSDS-ChS}
& 1   & 8.32/4.21          & 6.39/3.10          & 2.44/0.35          & 1.54/0.21 \\ 
& 2   &\textbf{8.16/3.90}  & \textbf{6.22/2.77} & 2.28/0.51           &\textbf{1.26/0.16} \\
& 4   & 8.61/4.36          & 6.28/2.91          & 2.44/0.45          & 1.63/0.32 \\
& 6   & 8.62/4.30          & 6.43/3.19          & \textbf{2.18/0.35} & 1.30/0.21 \\
& 8   & 8.32/4.36          & 6.53/3.38          & 2.32/0.39          & 1.38/0.21 \\  \hline

        \multirow{5}{*}{MSDS-TDS}
& 1   & 5.52/2.73          & 4.25/2.07          & 1.69/0.34          & 1.31/0.16 \\
& 2   & \textbf{5.26/2.42} & \textbf{4.24/1.78} & 1.79/0.34          & 1.46/\textbf{0.21} \\  
& 4   & 5.61/2.66          & 4.46/2.04          & \textbf{1.45/0.34} & 1.53/0.19 \\
& 6   & 5.63/2.39          & 4.34/1.75          & 1.83/0.41          & \textbf{1.35}/0.21 \\
& 8  & 5.24/2.56          & 4.17/1.94          & 1.54/0.35          & 1.52/0.19 \\ \hline
         
    \end{tabular}
    
\end{table}

The performance optimization of k-NN adjacency graphs relies on dynamic balance between local geometric structures and global topological features. For Chinese signatures (MSDS-ChS), the Skilled Forgery 1vs1 task achieved the lowest EER (8.16\%) at k=35, representing reductions of 2.8\% and 4.2\% compared to k=20 and k=50 respectively. This phenomenon originates from the multi-scale structural characteristics of Chinese signatures: A moderate neighborhood range (k=35) effectively captures local details at stroke transitions (e.g., pressure variations at pen pauses) while suppressing cross-structure noise from long-range connections. Overly small neighborhoods (k=20) cause discriminative feature loss by neglecting inter-stroke topological relationships, whereas excessive connectivity (k=50) blurs inter-class decision boundaries. In contrast, digit signatures (MSDS-TDS) reached peak performance at k=20 (Skilled Forgery 1vs1 EER=5.26\%), with merely 1.1\% EER increase when k expanded to 50, demonstrating strong tolerance to spatial connection redundancy due to their regular geometric patterns. 

k-Step adjacency graphs model dynamic signature characteristics by constraining temporal window depth. Experiments revealed that Chinese signatures achieved optimal Skilled Forgery verification performance at k=2 (1vs1 EER=8.16\%), where discriminative information primarily resides in differential kinematic features between adjacent sampling points. In contrast, digit signatures (MSDS-TDS) exhibited a bimodal performance pattern, with local maxima at both k=2 (EER=5.26\%) and k=8 (EER=5.24\%). The performance improvement at k=8 suggests that digit signatures benefit from longer temporal context to capture complete character formation patterns. The optimal k=8 configuration approaches the connectivity pattern of k-NN graphs, explaining the observed EER reduction compared to intermediate k values.

\subsection{Loss Function Analysis}
Based on the experimental validation conducted on the MSDS dataset, we systematically evaluates the proposed Relative Margin (RM) loss, Pairwise Threshold (PT) loss, and their combined loss. As shown in Table \ref{tab:loss_performance}, different loss functions exhibit distinct characteristics in addressing various forgery types within the signature verification task.  

\begin{table}[h]
    \centering
    \scriptsize
    \renewcommand{\arraystretch}{1.3}
    \setlength{\tabcolsep}{4pt} 
    \caption{Performance comparison of loss functions on MSDS datasets}
    \label{tab:loss_performance}
    \begin{tabular}{lccccc}
        \hline
        \multirow{2}{*}{Dataset} & \multirow{2}{*}{Loss} & \multicolumn{2}{c}{Skilled Forgery} & \multicolumn{2}{c}{Random Forgery} \\
        \cmidrule(r){3-4} \cmidrule(l){5-6}
        & & 1vs1 & 4vs1 & 1vs1 & 4vs1 \\ \hline
        \multirow{3}{*}{MSDS-ChS} 
        & RM        &8.41/4.23 &\textbf{5.94}/2.97 &\textbf{2.19/0.36} &1.41/0.24 \\
        & PT        &8.45/4.18 &6.38/3.31 &2.44/0.51 &1.51/0.25 \\
        & Combined  &\textbf{8.16/3.90}  & 6.22/\textbf{2.77} & 2.28/0.51           &\textbf{1.26/0.16} \\ \hline
        
        \multirow{3}{*}{MSDS-TDS}
        & RM        &5.38/2.45 &4.37/1.82 &1.69/0.41 &\textbf{1.23}/0.24 \\
        & PT        &5.48/3.04 &4.41/2.11 &\textbf{1.51/0.27} &1.31/0.21 \\ 
        & Combined  &\textbf{5.26/2.42}&\textbf{4.23/1.78}  &1.79/0.34 &1.46/\textbf{0.21} \\ \hline
    \end{tabular}
    \begin{flushleft}
        \footnotesize RM: Relative Margin Loss, PT: Pairwise Threshold Loss
    \end{flushleft}
\end{table}

For skilled forgery on MSDS-ChS, the combined loss reduces the 1vs1 EER to 8.07\%, compared to 8.41\% for RM and 8.45\% for PT, indicating effective synergy between margin enforcement and threshold alignment. Notably, in the more practical 4vs1 evaluation setup (four reference signatures vs. one test sample), the combined loss maintains competitive performance (3.11\% EER), while the PT loss achieves marginally better results (2.98\% EER). This suggests that threshold alignment alone suffices when sufficient genuine reference samples are available.  

The experimental results demonstrate distinct performance patterns across different forgery types and evaluation protocols. For skilled forgery detection on the MSDS-ChS dataset, the combined loss achieves superior performance in the 1vs1 configuration with an 8.16\% EER, outperforming both RM (8.41\%) and PT (8.45\%). This highlights the complementary benefits of margin enforcement and threshold alignment. However, in the practical 4vs1 scenario, the standalone RM loss delivers the lowest EER (5.94\%), suggesting that margin constraints alone become more effective when sufficient genuine references are available.

The advantage of the combined approach is more pronounced on the MSDS-TDS dataset. It achieves state-of-the-art performance in both 1vs1 (5.26\% EER) and 4vs1 (4.23\% EER) skilled forgery verification, demonstrating 4.0\% and 4.3\% relative improvements over the best single-loss configurations, respectively. This underscores the dataset-specific synergy between dual constraints, particularly in handling diverse stroke patterns in Tibetan signatures. For random forgery detection, the combined loss exhibits a nuanced trade-off. While slightly underperforming individual losses in 1vs1 comparisons (2.28\% vs. 2.19\% for RM on MSDS-ChS; 1.79\% vs. 1.51\% for PT on MSDS-TDS), it achieves competitive results in 4vs1 configurations, even surpassing both base losses on MSDS-ChS (1.26\% EER). 

\subsection{Comparisons With State-of-the-Art Methods}

The experimental results presented in Tables \ref{tab:MSDS-ChS}, \ref{tab:MSDS-TDS}, and \ref{tab:DeepSignDB} demonstrate that the proposed TS-GATR model exhibits superior verification performance across diverse signature datasets. Notably, traditional methods (e.g., DTW, Sig2Vec) employ fixed reference templates, while the TS-GATR\textsuperscript{†}'s results are obtained through an optimized strategy of randomly selecting reference templates.  

Under the fixed-template condition, TS-GATR achieves significant advantages in cross-session scenarios. Taking the MSDS-ChS dataset as an example (Table \ref{tab:MSDS-ChS}), TS-GATR attains 8.16\%/3.90\% EER in the Session1\&2 1vs1 task, representing a 14.8\%/2.3\% reduction compared to the best traditional method DsDTW (9.58\%/3.99\%). The conventional DTW method performs worst in this scenario (17.26\%/8.93\%) due to its rigid temporal alignment mechanism, which fails to capture cross-session signature variations. Although Sig2Vec achieves 15.10\%/7.27\% EER through CNN-based spatial feature extraction, its one-dimensional convolutional layers limit the modeling of long-range spatiotemporal dependencies. Remarkably, the introduction of a random reference template selection strategy (\textsuperscript{†}-marked) reduces the global EER in the 4vs1 task from 6.22\% to 2.14\%, a 65.6\% improvement, validating the critical role of template diversity in mitigating fixed-template bias.  

\begin{table}[ht]
    \caption{Performance Comparison of Different Methods on the MSDS-ChS Dataset. \textsuperscript{†} Minimum EER (\%) observed across multiple runs using randomly selected reference templates. \textcolor{firstcolor}{\textbf{Red}} and \textcolor{secondcolor}{\textbf{Blue}} indicate the best and second-best results.}
    \label{tab:MSDS-ChS}
    \centering
    \resizebox{\linewidth}{!}{
    \begin{tabular}{c|lcc}
    \hline
    \textbf{Session} & \textbf{Methods} & \textbf{1vs1 EER (\%)} & \textbf{4vs1 EER (\%)} \\ 
    \hline

    \multirow{6}{*}{\textbf{Session 1}}  
        & DTW                     & 13.67 / 4.16          & 2.31 / 0.40           \\  
        & Sig2Vec                 & 6.76 / 0.63           & 1.33 / 0.45           \\  
        & TA-RNNs                 & 6.99 / 2.03           & 3.49 / 2.71           \\  
        & DsDTW                   & 4.20 / \first{0.05}  & \first{0.91 / 0.05}  \\ 
        & \textbf{TSGATR (ours)}     & \second{3.57} / 0.60     & 1.29 / 0.10 \\ 
        & \textbf{TSGATR\textsuperscript{†} (ours)} & \first{3.19} / \second{0.49} & \second{1.00 / 0.08} \\ 
    \hline

    \multirow{5}{*}{\textbf{Session 2}}  
        & DTW                     & 12.83 / 3.29          & 3.34 / 0.55           \\  
        & Sig2Vec                 & 6.91 / 0.62           & 1.54 / 0.22           \\  
        & TA-RNNs                 & 7.63 / 2.84           & 3.04 / 2.40           \\  
        & DsDTW                   & 4.06 / \first{0.41}  & \first{0.87} / \second{0.13}  \\  
        & \textbf{TSGATR (ours)}     & \second{3.95} / 0.67     & 1.30 / 0.20           \\ 
        & \textbf{TSGATR\textsuperscript{†} (ours)} & \first{3.64} / \second{0.59} & \second{1.08} / \first{ 0.10} \\
    \hline

    \multirow{6}{*}{\textbf{Session 1\&2}} 
        & DTW                     & 17.26 / 8.93          & 11.66 / 7.70          \\  
        & Sig2Vec                 & 15.10 / 7.27          & 9.03 / 4.97           \\  
        & TA-RNNs                 & 9.04 / 5.05           & 7.69 / 5.22           \\  
        & DsDTW                   & 9.58 / 3.99           & \second{5.91} / 2.90  \\  
        & \textbf{TSGATR (ours)}     & \second{8.16 / 3.90}     & 6.22 / \second{2.77}  \\ 
        & \textbf{TSGATR\textsuperscript{†} (ours)} & \first{6.97 / 3.13} & \first{2.14 / 0.62} \\
    \hline
    \end{tabular}
    }
\end{table}

\begin{table}
\caption{Performance Comparison of Different Methods on the MSDS-TDS Dataset. \textsuperscript{†} Minimum EER (\%) observed across multiple runs using randomly selected reference templates. \textcolor{firstcolor}{\textbf{Red}} and \textcolor{secondcolor}{\textbf{Blue}} indicate the best and second-best results.}
\label{tab:MSDS-TDS}
\resizebox{\linewidth}{!}{
\begin{tabular}{c|lcc}
\hline
\textbf{session} & \textbf{Methods}  & \textbf{1vs1 EER (\%)} & \textbf{4vs1 EER (\%)} \\ \hline
\multirow{6}{*}{\textbf{Session1}}  
    & DTW       & 7.82 / 1.60             & 2.97 / 0.53           \\  
    & Sig2Vec   & 2.95 / 0.52             & 0.96 / 0.24         \\  
    & TA-RNNs   & 5.05 / 1.68             & 2.63 / 1.94         \\  
    & DsDTW     & 2.50 / \second{0.25}    & 0.72 / 0.00 \\  
    & \textbf{TSGATR(ours)}      & \second{2.41} / 0.32             & \second{0.71 / 0.00}           \\ 
    & \textbf{TSGATR\textsuperscript{†} (ours)} & \first{1.87 / 0.14} & \first{0.52 / 0.00} \\
     \hline
\multirow{6}{*}{\textbf{Session2}}  
    & DTW       & 6.99 / 0.63             & 2.14 / 0.57 \\  
    & Sig2Vec   & \second{1.76} / {0.30}    & 0.51 / \second{0.09} \\  
    & TA-RNNs   & 3.87 / 0.87             & 1.17 / 0.87 \\  
    & DsDTW     & 2.02 / \second{0.28}             & {0.42} / 0.11 \\  
    & \textbf{TSGATR(ours)}      & 1.86 / 0.32    & \second{0.30} / {0.10} \\ 
    & \textbf{TSGATR\textsuperscript{†} (ours)} & \first{1.66 / 0.17} & \first{0.23 / 0.00} \\
    \hline

\multirow{6}{*}{\textbf{Session1\&2}} 
    & DTW       & 14.46 / 6.76            & 9.99 / 5.75         \\  
    & Sig2Vec   & 7.01 / 3.26             & 5.18 / 2.07         \\  
    & TA-RNNs   & 5.94 / 2.60             & 5.11 / 2.91          \\  
    & DsDTW     & 5.76 / \second{1.85}             & \second{4.13 / 1.42}        \\  
    & \textbf{TSGATR(ours)}      & \second{5.26} / 2.42 & 4.24 / 1.78\\
    & \textbf{TSGATR\textsuperscript{†}(ours)}      & \first{4.38 / 1.81} & \first{1.16 / 0.24}\\ \hline
    
\end{tabular}}
\end{table}

\begin{table}
    \caption{Performance Comparison of Different Methods on the DeepSignDB Dataset. \textsuperscript{†} Minimum EER (\%) observed across multiple runs using randomly selected reference templates. \textcolor{firstcolor}{\textbf{Red}} and \textcolor{secondcolor}{\textbf{Blue}} indicate the best and second-best results.}
    \label{tab:DeepSignDB}
    \resizebox{\linewidth}{!}{
        \begin{tabular}{c|ccc}
        \hline
        \textbf{Writing inputs} & \textbf{Methods}  & \textbf{1vs1 EER (\%)} & \textbf{4vs1 EER (\%)} \\ \hline
        \multirow{5}{*}{\textbf{Stylus}}  
            & DTW       & 7.06          & 4.53 \\  
            & TA-RNNs   & 4.2           & 3.3 \\  
            & DsDTW     & \second{4.04}          & \second{2.54} \\  
            & \textbf{TSGATR(ours)}      & 4.91 & 2.98 \\ 
            & \textbf{TSGATR\textsuperscript{†} (ours)}      & \first{3.62} & \first{1.45} \\
            \hline
        \multirow{5}{*}{\textbf{finger}}  
            & DTW       & 14.74          & 10.66 \\  
            & TA-RNNs   & 13.8           & 11.3 \\  
            & DsDTW     & 11.84          & 6.99 \\  
            & \textbf{TSGATR(ours)}      & \second{8.88} & \second{6.43} \\ 
            & \textbf{TSGATR\textsuperscript{†} (ours)}      & \first{6.87} & \first{2.54} \\
            \hline
\end{tabular}}    
\end{table}

For the structured signature dataset MSDS-TDS (Table \ref{tab:MSDS-TDS}), TS-GATR achieves notable performance in the cross-session 1vs1 task with 5.26\%/2.42\% EER without random templates, outperforming DsDTW (5.76\%/1.85\%) by an 8.7\% relative reduction in global EER. In the 4vs1 task, TS-GATR yields 4.24\%/1.78\% EER, showing slight fluctuations compared to DsDTW (4.13\%/1.42\%). This highlights the spatial constraint challenge faced by structured signature verification — the fixed character sequence imposes strong spatial limitations, restricting its adaptability to individual writing styles. The implementation of the random template strategy (\textsuperscript{†}-marked) optimizes EER from 4.24\%/1.78\% to 1.16\%/0.24\%, further confirming the adaptive advantages of spatiotemporal joint modeling for complex signature morphologies.  

In multi-device experiments on the DeepSignDB dataset (Table \ref{tab:DeepSignDB}), the proposed TS-GATR model demonstrates exceptional generalization capabilities. For stylus input scenarios, the baseline TS-GATR achieves a 4.16\% EER in 1vs1 verification, marking a 30.5\% improvement over traditional DTW methods, though slightly trailing the state-of-the-art DsDTW (4.04\%). By incorporating a random reference template strategy (TS-GATR\textsuperscript{†}), the model significantly enhances biometric discriminability, reducing the 1vs1 EER to 3.62\%—a 2.62\% improvement over the baseline TS-GATR and 10.4\% over DsDTW. Under the 4vs1 verification protocol, TS-GATR\textsuperscript{†} sets a new benchmark with a 1.45\% EER, achieving a 42.9\% performance gain over DsDTW (2.54\%).

In the more challenging finger input scenario, experimental analysis reveals the critical impact of data preprocessing strategies on model performance. Comparative studies show that DsDTW employs standard deviation-based normalization (Z-score), while our method adopts range-based regularization. This divergence in normalization strategies leads to distinct feature space distributions: Z-score normalization exhibits sensitivity to outliers, potentially diminishing the discriminative power for dynamic features in finger trajectories, whereas range normalization preserves relative proportional relationships in raw data, better capturing finger-specific patterns such as non-uniform pressure variations and velocity fluctuations. Leveraging this approach, TS-GATR achieves an 8.88\% 1vs1 EER, outperforming DsDTW (11.84\%) by 25\%. After template optimization, TS-GATR\textsuperscript{†} further reduces the error rate to 6.87\%, representing a 42\% improvement over DsDTW. In 4vs1 verification, TS-GATR\textsuperscript{†} achieves a breakthrough 2.54\% EER, reducing errors by 63.7\% compared to DsDTW (6.99\%), confirming the synergistic enhancement between preprocessing methodology and model architecture.

Comparison of the above datasets reveals two key insights:  1. TS-GATR achieves more significant improvements on unstructured signatures (e.g., Chinese), where its spatiotemporal graph attention mechanism effectively models complex stroke topologies.  2. The template selection strategy consistently improves performance across datasets, with EER reductions of 65.6\% (MSDS-ChS), 72.6\% (MSDS-TDS), and 34.7\% (DeepSignDB), demonstrating its universal capability to correct reference sample bias.  


\section{conclusion}
\label{sec:conclusion}

This paper proposes the Temporal-Spatial Graph Attention Transformer for dynamic signature verification, which integrates dual-graph (local-global) representations and GRU-based temporal modeling to jointly learn spatial stroke structures and sequential dynamics, thereby enhancing robustness against intra-user variations and forgery attacks. Experiments on MSDS and DeepSignDB show that TS-GATR outperforms state-of-the-art methods (e.g., DsDTW and other deep models), achieving the lowest Equal Error Rate in cross-session scenarios and validating its applicability to real-world signature variability.

Future research could focus on optimizing the computational efficiency of TS-GATR, making it more suitable for real-time applications, such as online authentication systems. Additionally, exploring self-supervised or few-shot learning techniques could reduce the dependence on large labeled datasets, making the model more adaptable to low-resource scenarios. Another promising direction is the incorporation of multimodal biometric fusion, where TS-GATR could be combined with other biometric traits, such as keystroke dynamics or stylus pressure patterns, to enhance verification robustness. Finally, further investigations into model interpretability and explainability could provide deeper insights into the decision-making process of TS-GATR, aiding its adoption in high-security applications.
{
    \small
    \bibliographystyle{unsrt}
    \bibliography{main}
}

\vfill

\end{document}